\documentclass{article}

\usepackage{arxiv}

\usepackage[utf8]{inputenc} % allow utf-8 input
\usepackage[T1]{fontenc}    % use 8-bit T1 fonts
\usepackage{hyperref}       % hyperlinks
\usepackage{url}            % simple URL typesetting
\usepackage{booktabs}       % professional-quality tables
\usepackage{amsfonts}       % blackboard math symbols
\usepackage{nicefrac}       % compact symbols for 1/2, etc.
\usepackage{microtype}      % microtypography
\usepackage{lipsum}		% Can be removed after putting your text content
\usepackage{graphicx}
\usepackage{natbib}
\usepackage{doi}

\usepackage{times}
\usepackage{latexsym}

\usepackage[T1]{fontenc}
\usepackage{amsmath,amssymb,amsfonts}
\usepackage{algorithm}
\usepackage{algorithmic}

\usepackage{graphicx}
\usepackage{textcomp}
\usepackage{adjustbox}
\usepackage{multirow}

\usepackage{microtype}

\usepackage{inconsolata}

\title{FedPC: Federated Learning for Language Generation with Personal and Context Preference Embeddings}

\author{ {\hspace{1mm}Andrew Silva}\thanks{Denotes equal contribution} \\
	School of Interactive Computing\\
	Georgia Institute of Technology\\
	Atlanta, GA\\
	\texttt{andrew.silva@gatech.edu} \\
	\And
	{\hspace{1mm}Pradyumna Tambwekar$^*$} \\
	School of Interactive Computing\\
	Georgia Institute of Technology\\
	Atlanta, GA\\
	\texttt{ptambwekar3@gatech.edu} \\
	\And
	{\hspace{1mm}Matthew Gombolay} \\
	School of Interactive Computing\\
	Georgia Institute of Technology\\
	Atlanta, GA\\
	\texttt{matthew.gombolay@cc.gatech.edu} \\
}

\renewcommand{\shorttitle}{FedPC}

\hypersetup{
pdftitle={FedPC: Federated Learning for Language Generation with Personal and Context Preference Embeddings},
pdfsubject={q-bio.NC, q-bio.QM},
pdfauthor={Andrew Silva$^*$, Pradyumna Tambwekar$^*$, Matthew Gombolay},
pdfkeywords={Natural Language Processing, Federated Learning, Personalization},
}

\begin{document}
\maketitle

\begin{abstract}
Federated learning is a training paradigm that learns from multiple distributed users without aggregating data on a centralized server. Such a paradigm promises the ability to deploy machine-learning at-scale to a diverse population of end-users without first collecting a large, labeled dataset for all possible tasks.
As federated learning typically averages learning updates across a decentralized population, there is a growing need for personalization of federated learning systems (i.e conversational agents must be able to personalize to a \textit{specific} user's preferences).
In this work, we propose a new direction for personalization research within federated learning, leveraging both personal embeddings and shared context embeddings.
% , providing a more efficient and effective path to personalization.
% By learning a global language model and solely relying on preference and context embeddings for personalization, the federated learning model can continually improve with all users' data. 
% Our model learns to separate out preferences into personal and contextual embeddings, personalizing to diverse users while pooling data for shared context embeddings. 
We also present an approach to predict these ``preference'' embeddings, enabling personalization without backpropagation. 
% We show that these generated embeddings exhibit comparable performance in terms of perplexity to embeddings learned via backpropagation. 
Compared to state-of-the-art personalization baselines, our approach achieves a 50\% improvement in test-time perplexity using 0.001\% of the memory required by baseline approaches, and achieving greater sample- and compute-efficiency.
\end{abstract}

% keywords can be removed

\section{Introduction}
As conversational agents and dialog systems are deployed to real-world scenarios, these systems require data-efficient personalization paradigms such that language systems such as conversational agents can be effectively adapted on-device.
The benefits of on-device optimization are two-fold; (1) Swift adaptation of model-behavior based on human-interactions \citep{dudy2021refocusing}, (2) Privacy protection by means of retaining all data related to the user on-device \citep{li2020secure}.
% agents can quickly adapt their behavior based on their interactions with humans\citep{dudy2021refocusing}, (2) agents can protect the user's privacy by means of not having to send the user's data back to the server \citep{li2020secure}. 
One of the prevailing paradigms for learning from and engaging with end-users is \textit{federated learning}.
Federated learning is an inherently decentralized learning paradigm that assumes no access to a large labeled dataset and instead leverages averaged parameter updates across all users of the system \citep{mcmahan2017communication}. Such averaged updates invariably dilute individual preferences or deviations from the mean, resulting in a model that works well for the average user while failing to appropriately capture under-represented preferences or sub-groups within the data. 
% Users with such preferences are left with a model that does not work well for their particular data distribution, as the model is fit to the average persona and neglects individual preferences. 
In this work, we present a novel approach (FedPC) to personalizing federated learning with personal and context embeddings (collectively called ``preference embeddings''), adapting more efficiently and effectively than prior work with respect to both data and compute on-device.

\begin{figure}[t]
    \centering
    \includegraphics[width=0.7\linewidth]{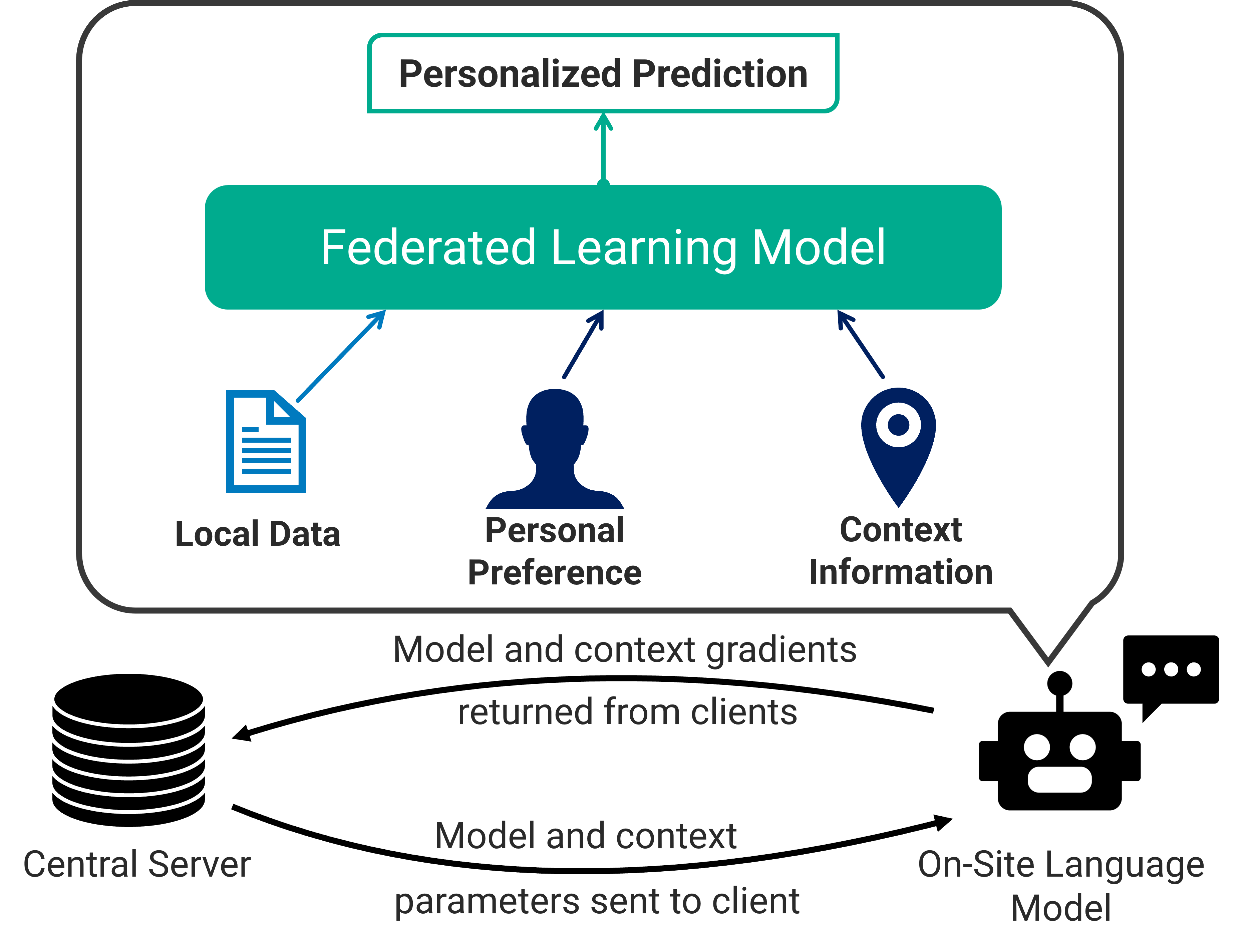}
    \caption{Overview of our personalized federated learning setup, FedPC. Language models within client devices, such as individual agents deployed to communicate with people at hospitals, homes, or construction sites, pull down global model parameters and context embeddings. Local, on-device data is then paired with both personal and context embeddings to produce personalized predictions with global model parameters.}
%    \todo{@Andrew, do you know what Matthew means by ``weird box artifact''?}}
    \label{fig:pipeline}
\end{figure}

% Within federated learning, we do not have access to a large, labeled dataset, but instead we leverage a distributed set of small datasets from individual user devices, such as conversational agents for robots deployed to hospitals, homes, or other domains with evolving dynamics and regular human interactions. 
% % These clients first pull down a large neural network from a centralized server. Clients then make local updates to the network and send their gradient information back to a central server, which then aggregates updates from all users into one global update. The global update is applied to a global, server-side model. After this update, the global model is pushed down to all clients, and the process repeats. 
% Under the federated learning paradigm, it is possible to learn supervised models over personal or private data, such as medical images \citep{rudovic2021personalized}, without aggregating all private data in one central location. 
% % However, as the model is continually applying \textit{averaged} gradient updates, the model fails to capture individual preferences for users that are atypical. 
% While personalization has been explored in centralized setups \citep{song-etal-2020-learning,song-etal-2021-bob}, such approaches assume access to a centralized dataset. 
% Several approaches have also been proposed to extend personalization to federated learning, though such methods often assume that there exists only a small number of users, and that users have little useful overlap in their underlying data distributions. 
 
We leverage the insight that a client's data distribution is informed by both individual preferences and additional contextual information. 
% A user's preference or personas should be reflected in the responses of an interactive language agent to enable smooth interactions. 
 For example, while each user may have their own \textit{individual} style, there may be more general \textit{population-wide} trends that inform the style of personalized predictions (e.g., dialogue assistants helping patients with cognitive disorders, whereby agents can personalize to individual patients and broader condition-wide trends).  
 While individual preferences may be unique to each client (e.g. a user's taste or affect), we can more accurately personalize to client preferences with the addition of context, as shared-context parameters carry beneficial stylistic information across clients \citep{dudy2021refocusing,jones1999automatic}. Stylistic or situational context provides additional information to curate relevant language outputs that can be shared across users. 
 % Context may include critical additional information that is not specific to the user nor explicitly part of the input sample, such as the time of day, location of the device, or domain. By leveraging added context, we can more accurately personalize to client preferences, as shared-context parameters carry beneficial stylistic information across clients \citep{dudy2021refocusing,jones1999automatic}. 
%  Furthermore, incorporating stylistic preferences and personas into generations can affect the perception towards the ``correctness'' of user-grounded generations.
%and more rapidly personalize to new clients.

In this work, we contribute a new approach to personalized federated learning that is both easier to learn and more effective than prior work, and investigate the utility of personalization via individual preferences and contexts.
% Our work provides an approach to learn personal and context embeddings from a user's local data, both of which are integrated into a pre-trained DistilGPT2 model, to facilitate personalized language generation. 
While prior language generation approaches have developed personal or persona-based generative systems \citep{wu2021personalized,zhang2018personalizing} or context-based generative systems \citep{cheng2019dynamic, lin2019moel} individually, none have combined them to personalize outputs in a low-data setting under stylized preferences. We show that our approach is more sample-efficient than state-of-the-art baselines, while requiring less time to train. We additionally present an inference-only version of our approach, personalizing without backpropagation for new users. Finally, we directly test the potential for personalization with users who have been held-out from training (i.e., testing with new users). An overview of our approach is given in Figure \ref{fig:pipeline}.

%Our approach affords improved generalization by individually modelling a user's personal and context-related preferences.
% , enabling us to modulate a user's predictions based on a given context. 

% We show that within a federated learning setting, where user-information is retained on-device and each user has a small local dataset, FedPC is more sample-efficient and data-efficient than prior state-of-the-art baselines.
% Furthermore, we provide extensive analysis of the performance of FedPC on held-out datasets, where the model is tested on users whose data has been held-out during global optimization.
% Such experiments on data efficiency for \textit{new} users are often missing in the related literature on personalized language generation, despite new-user generalization being a critical component of any such framework.
% Finally, we develop a first-of-its-kind approach to generate preference embeddings through inference alone, without any on-device backpropagation.
% Our approach can generate meaningful personal and context embeddings using a single sample while still exhibiting comparable performance to baselines, thus providing a highly-efficient means of performing on-device personalization. 
% An overview of our approach is given in Figure \ref{fig:pipeline}.

\section{Related Work}
Federated learning enables machine-learning at-scale to a diverse population of end-users without first collecting a large, labeled dataset for all possible tasks. After the introduction of \textit{federated averaging} \citep{mcmahan2017communication}, focus has shifted to different ways of personalizing to individual users. 
Prior personalization approaches for federated learning have typically involved learning personal network heads and a shared global encoder (i.e., ``split-learning'' approaches \citep{gupta2018split}), or learning a separate local model from a global initialization (i.e., a ``meta-learning'' approach \citep{finn2017maml,nichol2018reptile}).
% While there have been several personalization approaches for federated learning, such approaches typically involve  

% In our work, we propose a new approach to personalized federated learning through on-device personal embeddings with a shared global model, enabling personalization with drastically reduced computational overhead and data requirements.

\paragraph{Learning Personal Model Heads}
\label{subsec:personal-heads}
The most prevalent approach to personalization in federated learning is through personalized model heads. Such approaches share gradient information to learn a global feature encoder, but retain user-specific classification-head gradients on-device. Approaches such as FedRep \citep{collins2021fedrep} solely separate out local and global gradients, while other methods such as PFedMe \citep{dinh2020pfedme} enforce constraints on model-divergence (such as via FedProx\citep{li2020federated}). Other approaches, such as FedMD \citep{li2019fedmd}, enable clients to adopt any desired architecture, sharing a common backbone but allowing for completely divergent model heads \citep{arivazhagan2019federated,kim2021spatio,rudovic2021personalized,paulik2021federated}. Finally, there has recently been increased effort on identifying clusters of related users to share model heads, such as with K-Means clustering in PFedKM \citep{tang2021pfedkm} or through clustered personal embeddings in FedEmbed \citep{silva2022fedembed}. Notably, there is no prior work which learns both personal \textit{and} contextual model heads for personalization within federated learning.
%, \change{which applies hard K-Means clustering to client gradients}.
% However, such work may lead to privacy violations for individual users \citep{chuanxin2020federated,li2020secure}, \change{as user-data is shared across client devices}.

\paragraph{Meta-Learning Global Models}
\label{subsec:meta-learned}
An alternate approach to personalizing federated learning models is through the adoption of meta-learning \citep{jiang2019improving, fallah2020personalized}, for learning a global model prior to fine-tuning on client-data. 
% As shown by prior work \citep{jiang2019improving,fallah2020personalized}, federated learning can be viewed as a special case of meta-learning. 
%As such, recent research has viewed global models as meta-learned ``average'' models of all user behavior. 
After cloning the global model as an initialization from all client's updates, local, client-side models are permitted to diverge and fine-tune to a user's individual preferences or data distribution \citep{fallah2020personalized,deng2020apfl,hanzely2020federated,hanzely2020lower, lin2019personalizing}. However, computing and applying gradients for a full model often requires too much time, power, and memory. As such, expensive full-model gradients can often only be computed and applied when a device is not actively in-use. As in the split-learning literature, there are not meta-learning approaches for disentangling personal and contextual preferences within personalized federated learning.

% However, because client devices may lack the data to adequately fine-tune the full network, such training setups often involve blending the output of the global and local models for final output predictions.

\paragraph{Learning with Personal Embeddings}
Our work leverages the insight that personal preferences can be represented using a personalized embedding, allowing the model to condition output predictions on personal preferences without requiring completely re-trained classification heads or networks. Personal embeddings have been used in prior work to capture an individual's ``style,'' often in imitation learning settings \citep{tamar2018imitation,hsiao2019learning,paleja2020interpretable,schrum2022mind}. 
Treating personal embeddings as neural network parameters that are updated on-device, these approaches learn to embed preferences and condition network output over both input data and preference embeddings. Most closely related to our work are FedNLG \citep{lu2021federated}, which predicts ``persona'' parameters for users, and the Global+ model in FedEmbed~\citep{silva2022fedembed}, which learns a personal embedding for each user. 
However, FedNLG requires access to a user's entire history of language and demographic data in order to produce a ``persona'' for each user, informing the generation of a ``persona'' embedding, and Global+ incorporates supervised style feedback. Prior embedding-based approaches solely learn \textit{personal} embeddings, neglecting stylization through context. In our work, we explore the utility of incorporating context in addition to personal preferences, and all preference embeddings are updated solely via a self-supervised language-modeling loss. % through a data-efficient procedure, and our preference embeddings are updated solely via a self-supervised language-modeling loss.

%\todo{@Andrew, can we remove this last sentence? If not, we probably need to cut a few lines in this section.}
% of incorporating context in addition to personal preferences .

\paragraph{Personalization in Language}
Personalization for language generation systems seeks to produce grounded systems that can efficiently adapt to end-user needs \citep{yang2021towards, dudy2021refocusing}.
% , and embeddings have become a popular avenue of personalizing language models. 
One such approach to personalization is by learning a ``persona'' for each user and conditioning the language model on the embeddings or representation for the persona via a memory network \citep{zhang2018personalizing, wu2021personalized, lu2021federated}. 
``Personas'' are generally short sequences of 5-6 sentences which contain information about an individual such as ``I have blonde hair'' or ``My mom is a doctor.''
Similar approaches leverage Bayesian inference methods to infer context\citep{majumder2020like} or persona \citep{kim2020will}, and then condition the language generation on the inferred context. 
However such approaches involve collecting and maintaining user-profiles on a central server which may violate user-confidentiality. 
Alternate approaches seek bypass this issue by enabling dynamic speaker modeling through context-based fine-tuning rather than conditioning on profile information \citep{cheng2019dynamic, li-liang-2021-prefix}. 
%These methods learn contextual speaker embeddings or model-weights (prefixes) by fine-tuning the model on conversation history of a given user.
FedPC leverages a similar design to dynamically learn personal and context embeddings through data from small datasets for a given user, while also preserving user-confidentiality via federated learning. 
% Dynamic Speaker model for conversational context - \citep{cheng2019dynamic}. Learns speaker information from conversation history rather than private information about the user. Useful for learning embeddings for new users

% \citep{wu2021personalized} - Personalization through profile and previous message information. Profile (age, gender, interests), message (prior messages related to query). Both profile and message information is encoded through separate attention mechanisms to create an embedding for the user, which is then forwarded to the decoder. Requires preexisting profile information. 

% \citep{zhang2018personalizing} - First work on personalization through profile information. Encode profile information through memory network and incorporate with LSTM encoder. Privacy issues. Profile information

% Bayesian inference of context \citep{majumder2020like} or persona choice\citep{kim2020will}

% Prefix-tuning - \citep{li-liang-2021-prefix}

FedPC represents a new direction in personalized federated learning research, enabling personal and stylized language generation with a fraction of the memory, data, and compute costs of prior approaches without requiring access to pre-made personal profiles.
\section{Approach}
\begin{figure}[t]
    \centering
    \includegraphics[width=0.7\linewidth]{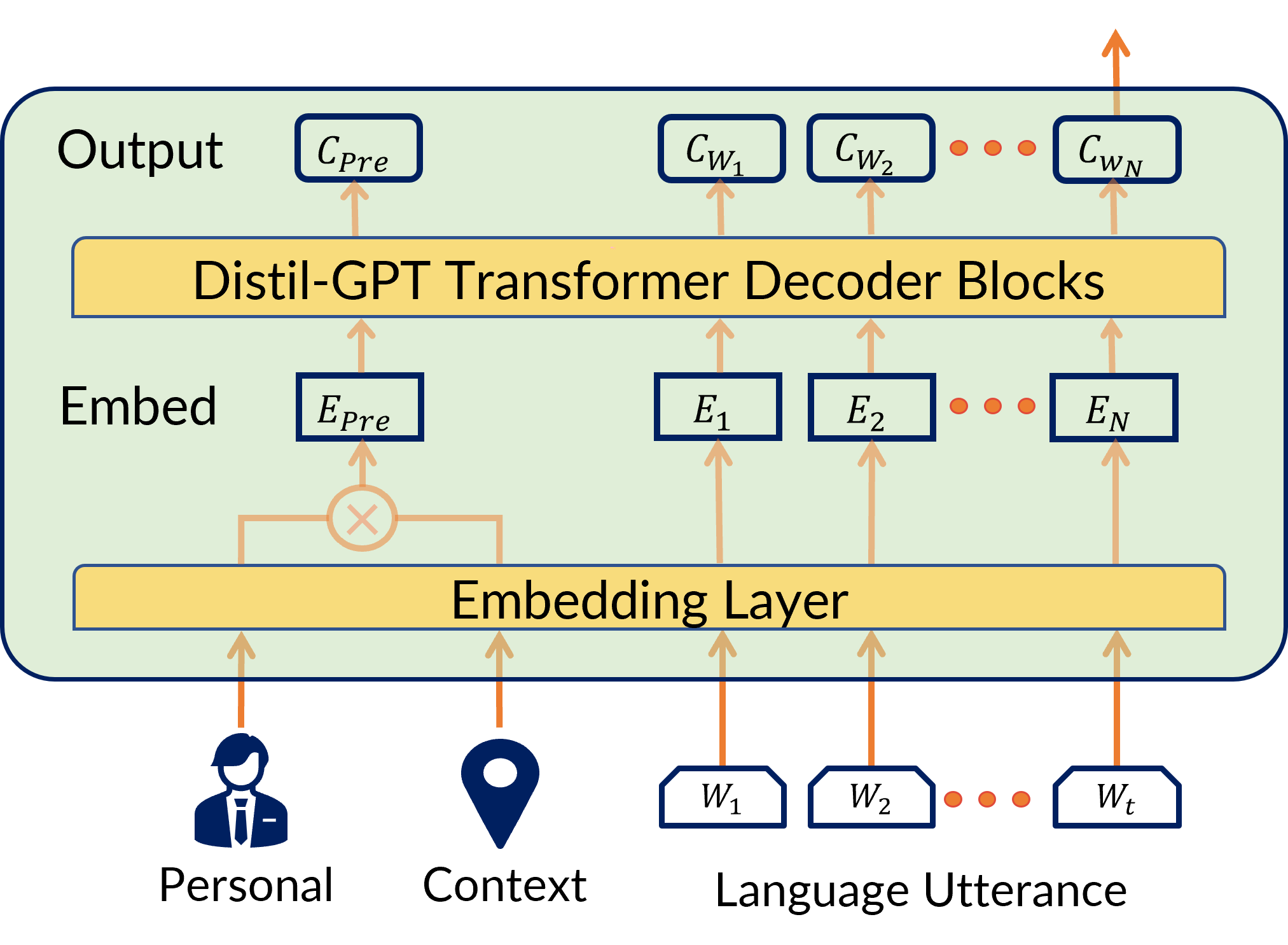}
    \caption{The FedPC model architecture. Input data, such as on-device conversation data for a user, is passed into the language model in addition to personal and context labels specifying user's preference. The personal and context labels are embedded through a preference embedding layer to produce a single preference embedding. 
    % The personal and context embeddings generated are then combined into a single ``preference'' embedding via an element-wise dot product.
    This preference embedding is combined with the word embeddings for the input sequence and passed into the DistilGPT2 model to predict the next word.}  
    \label{fig:architecture}
\end{figure}

%In this section, we demonstrate initial results of federated learning with personalized embeddings, and elucidate the benefits of personalization for on-device learning. 
In this section, we present our novel approach to personalization in federated learning with FedPC.
FedPC produces personal and contextual preference embeddings either via backpropagation (i.e., learning preference embeddings), or by inference (i.e., predicting preference embeddings).
% We present two approaches to obtaining preference embeddings for individual users or for shared contexts.
% We note that any federated learning paradigm could be exchanged for FedAvg or FedProx, 
% We note that the specifics for learning a global feature encoder are not relevant to our primary contribution, as our approach focuses on the addition of personalized embeddings for federated learning.
A visual overview of our federated learning architecture is in Figure \ref{fig:architecture}, and a step-by-step walk-through of our training algorithm is given in Algorithm \ref{alg:training_loop}.

\subsection{Personalization via Embeddings}
Personalization in FedPC is achieved entirely through preference embeddings. Every input sample (e.g., an incomplete sentence) is accompanied by both a personal preference embedding, representing the user, and a contextual preference embedding, representing the context or style of the prediction.
These two embeddings are combined via an element-wise multiplication to produce a single preference embedding that accompanies the input sample.
By leveraging both personal and context embeddings, FedPC considers the individual user \textit{and} the broader context of an utterance, enabling personal, stylized prediction.

In the language-modeling domain, the unified preference embedding is prepended to the input utterance, providing a prefix for the model to consider \citep{li-liang-2021-prefix}.
The model then predicts the next token of the utterance, and a language-modeling loss is calculated by comparing the prediction to the user's actual next token. The next token is then appended to the sequence, and preference embeddings are again prepended to the new input sequence, and the process repeats. After completing a full utterance, preference embeddings may be updated, either through backpropagation or by using an embedding-generator to predict new personal and contextual embeddings for the client.

\subsection{Federated Learning Algorithm}
% Here, we present FedPC with differentiable, learned preference embeddings.
To begin, all clients initialize their own personal embedding on-device, and the server initializes a set of $C$ context vectors for each relevant setting given the target task.
% (e.g., expecting a ``professional'' and ``casual'' context for next-word prediction on smartphones).
We additionally assume that all data points on a client device have an associated context, $c$, being derived from the contextual information of the client device when the data point was captured (e.g., time of day, location, etc.).

Training begins by distributing all the requisite information to client-devices. 
Client devices pull down the global model parameters, $\theta$, and the global context embedding parameters, $\phi$, making local copies, $\theta_d$ and $\phi_d$ (line 6).
Unlike the global model parameters and context embeddings, the personal embeddings, $\psi_d$ do not need to be copied from the server as they are kept on client-devices.

% Algorithm~\ref{alg:training_loop} shows our approach with embeddings learnt via backpropagation, however our approach also allows for embeddings to be generated via an embedding-generator. 

Client devices then take $K$ gradient steps using their own on-device data, where each input sample is paired with the client's on-device embedding, $\psi_d$, and the context embedding for the particular sample, $\phi_{d,c}$, assuming the data point was drawn under context $c \in C$.
Gradients are calculated using a language-modeling objective, though any objective could theoretically be applied.
If preference embeddings are being generated via forward-propagation rather than learned via backpropagation, contextual and personal preference embeddings will also be predicted by an embedding-generator at this stage (note: the parameters of the embedding-generator are shared globally, being a part of $\theta$).%substituted in as the task for the global model.

Gradients are applied to the shared-model parameters, $\theta_d$, and are then used to update preference embeddings (line 9).
If preference embeddings are being predicted, these gradient steps are also applied to the shared embedding-generator, and preference embeddings (i.e., context embeddings $\phi_d$ and personal embeddings $\psi_d$) are overwritten with their latest predicted values (lines 10-11).
If preference embeddings are being learned via backpropagation, gradient steps are applied to $\phi_d$ and $\psi_d$ using Equation \ref{eqn:backprop_embeds} (lines 10-11).
% \todo{@Matthew, We've used the term preference embeddings to represent both personal and context embeddings. We've added to the first usage of the term ''preference embeddings'' to specify that this means both sets of embeddings. Is there anything else you would like us to change to make this terminology clearer}

%Each gradient step is applied to $\theta_d$, $\phi_{d}$, and to the client's on-device embedding, $\psi_d$ (lines 11 - 13).

After $K$ steps, gradients for $\theta_d$ and $\phi_d$ are sent back to the server, while $\psi_d$ remains on-device (lines 13 - 15).
The server computes a single update for the global model and context embeddings by averaging across all clients (lines 17-18).
% The server aggregates gradients from all client devices, averaging them into a single update for the global model and context embeddings. 
The server applies the averaged update to $\theta$ and $\phi$, and the process repeats (lines 19-21).

\begin{algorithm}[t]
\caption{FedPC Training Loop}
\label{alg:training_loop}
\begin{algorithmic}[1]
\STATE {\bfseries Given:} Training objective $\mathcal{L}$, Client devices $D$, \# client steps, $K$, \# global steps, $N$
\STATE {\bfseries Initialize:} Global model $\theta$, Context embeds $\phi$  % context embeddings $\phi$
\STATE{\bfseries Initialize:} Personal embeddings on-device $\psi$
\FOR{$n \in N$}
    \FOR{$d \in D$}
        \STATE $\theta_d = \theta, \phi_d = \phi$
        \FOR{$k \in K$}
            \STATE Sample $B_d$ from user's on-device data
            \STATE $\theta_d \leftarrow \theta_d + \nabla_\theta\mathcal{L}(\theta_d, \phi_{d,c}, \psi_d, B_d)$ % // Fine-tune global model
            % \STATE $\phi_d \leftarrow \phi_d + \nabla_\phi \mathcal{L}(\theta_d, \phi_{d,c}, \psi_d, B_d)$ % // Fine-tune context embedding
            % \STATE $\psi_d \leftarrow \psi_d + \nabla_\psi \mathcal{L}(\theta_d, \phi_{d,c}, \psi_d, B_d)$ % // Update personal embedding
            \STATE $\phi_d \leftarrow \phi_{d} + \nabla_{\phi_d}$
            \STATE $\psi_d \leftarrow \psi_d + \nabla_{\psi_d}$

        \ENDFOR
        \STATE $\nabla_{\theta_d} \leftarrow \theta-\theta_d$ % // compute $\theta$ gradients
        \STATE $\nabla_{\phi_d} \leftarrow \phi - \phi_d$  % // compute $\phi$ gradients
        \STATE Return $\nabla_{\theta_d}$ and $\nabla_{\phi_d}$ to the server
    \ENDFOR
    \STATE $\nabla_\theta \leftarrow \frac{1}{D}  \sum_d^D \nabla_{\theta_d} $  % //  calculate average $\theta$ gradients
    \STATE $\nabla_\phi \leftarrow \frac{1}{D}  \sum_d^D \nabla_{\phi_d} $ % //  calculate average $\phi$ gradients
    \STATE $\theta \leftarrow \theta + \nabla_{\theta}$
    \STATE $\phi \leftarrow \phi + \nabla_{\phi}$
\ENDFOR
\end{algorithmic}
\end{algorithm}
\begin{equation}
    \label{eqn:backprop_embeds}
    \begin{aligned}
     &\phi_d = \phi_d + \nabla_\phi\mathcal{L}(\theta_d, \phi_{d,c}, \psi_d, B_d) \\
     &\psi_d = \psi_d + \nabla_\psi \mathcal{L}(\theta_d, \phi_{d,c}, \psi_d, B_d)
     \end{aligned}
\end{equation}
In a typical federated averaging deployment, client devices will pull down global parameters, fine-tune on local datasets, and then test on held-out, local data. 
With FedPC, the majority of the network's parameters, $\theta$, are frozen, reflecting a federated-learning setup with a more constrained computational budget when deploying large language models.
Using FedPC, clients pull down and subsequently freeze global parameters, $\theta$, and either generate preference embeddings from observation, or only compute and apply gradients to context embeddings, $\phi$, and their local personal embedding $\psi$.
Relying on forward-propagation calls rather than backpropagation, or by computing gradients over only these embeddings, we reduce the computational overhead of FedPC while preserving or even improving upon accuracy relative to fine-tuning an entire model.
When testing over local data, all updates to context embeddings $\nabla_\phi$ are not sent to the central server.
% , as there is no global-learning performed during evaluation. 
Rather, these gradients are directly applied to the context embeddings for the current user, and then discarded. When instantiating a new embedding for a previously unseen user, we set the user's embedding to the noisy-average of all known user embeddings.

% , thereby providing a warm-start to the new user.

\subsubsection{Generating Preference Embeddings}
% Rather than learn preference embeddings through backpropagation, 
% We also present an approach to \textit{predicting} preference embeddings, via on-device inference alone.
To generate embeddings, we adopt a similar procedure to HyperNetworks \citep{ha2016hypernetworks,shamsian2021personalized}, in which a neural network is trained to predict parameters of another network. In FedPC, an embedding-generator is trained to predict the parameters of preference embeddings (either personal or context).
To generate embeddings, we apply an additional transformer decoder block~\citep{vaswani2017attention}, that uses a randomly-initialized personal embedding and a known context embedding as the queries, along with the word embeddings for the utterance as the keys and values to update the given preference embeddings.
We utilize separate generators to predict the personal embedding, $\psi_d$, and the context embedding, $\phi_d$. 
Specific training details for the embedding-generator applied to language-modeling are given in the appendix.

% Given an utterance and a preference embedding, the embedding-generator applies multi-head self-attention, layer norm, and a multilayer-perceptron to produce a new preference embedding for the given utterance.
% This process is repeated for the personal embedding, $\psi_d$, and the context embedding, $\phi_d$. Specific training details for the embedding-generator applied to language-modeling are given in the appendix.

While the embedding-generator must be learned from scratch during training, this method of predicting preference embeddings allows us to generate personal embeddings for previously \textit{unseen} users when testing.
By predicting preference embeddings, we circumvent the need for expensive gradient calculation and on-device learning. Instead, new users can quickly reap the benefits of personalized predictions via a trained preference prediction module (i.e., the embedding generator), as opposed to conventional personalized federated learning methods that require slow and sample-inefficient on-device learning.

\section{Experiments}
We conduct several experiments to evaluate the sample efficiency, generalization, and runtime of our approach relative to baseline federated learning frameworks. In our experiments, we compare:
\begin{itemize}
    \item FedPC  -- Learning personal and context embeddings jointly with a global feature encoder, and performing local fine-tuning of personal and context embeddings on-device.
    \item FedPC (Frozen) -- As above but without local fine-tuning.
    % , thereby isolating the contributions of refined personal and context embeddings.
    \item FedPC (Generated) -- Learning an embedding generator and global feature encoder, and then using only generated embeddings at test-time.
    % , showcasing the ability of personalization in federated learning with \textit{no} on-device learning.
    \item Split-Learning -- Learning personal and context-specific model-heads jointly with a global feature encoder, and performing local fine-tuning of the personal and context-specific model heads on-device \citep{dinh2020pfedme,collins2021fedrep}.
    \item Meta-Learning -- Learning a single global model for all users and contexts, and fine-tuning the shared model-head on-device \citep{finn2017maml,nichol2018reptile}.
\end{itemize}

We conduct two sets of experiments to compare the above approaches on both sample efficiency and runtime efficiency. For the sample efficiency experiments, we present perplexity numbers for all methods across two versions of the dataset: known users and withheld users. For our known user experiments, all users are present in the training and testing set. For our withheld user experiments, a subset of users from each dataset is withheld entirely from training, and performance results are presented only for the held-out users. Perplexity is calculated over unseen utterances with the first three tokens of each utterance given as a prompt. Finally, we present qualitative results from our method, demonstrating the power of stylized generation for individual users.

All models are initialized with the DistilGPT2 pre-trained model from Huggingface \citep{wolf2019huggingface}, with all layers frozen. For Split-Learning and Meta-Learning, the last layer of the model is unfrozen. Training details are in the appendix.

%All layers of the model are frozen, and FedPC only backpropagates error to personal and context preference embeddings. For our Meta-Learning baseline, the last layer is unfrozen and all users jointly update this final output layer (note that there is no dedicated context head in this approach). Our Split-Learning baseline assigns a unique model head to each user and to each context, and each user only updates their own model head and the contexts that they use. All models are trained for 55 epochs over their training datasets using the Adam optimizer \citep{kingma2014adam} for global updates (learning rate = 1) and local updates (learning rate = 0.001). Each client (character or Reddit user) makes 10 local updates before passing their pooled gradient information back to the server.

% \begin{table}
% \normalsize
% \caption{Dataset Statistics}
% \label{table:datasets}
% % \begin{adjustbox}{width=1\textwidth}
% \centering
% \begin{tabular}{l|cccccccc}
% Dataset-Type & Users         & Utterances            & Contexts\\ %           & Utterances per user & Utterances per context & Contexts per user & Users per context             \\
% %\hline
% %Perplexity    & 72.54\% $\pm$ 1.31\% & 93.90\% $\pm$ .20\% & 93.52\% $\pm$ .45\% & 74.52\% $\pm$ 2.68\% \\
% \hline \hline
% Reddit & 326 & 30260 & 57\\ %& 93 $\pm$ 61 & 530 $\pm$ 365 & 1.83 $\pm$ 1.09 & 10.44 $\pm$ 6.28 \\
% TV Shows & 19 & 60650 & 2 %& 3369 $\pm$ 3342 & 30325 $\pm$ 17896 & 1.0 $\pm$ 0 & 9 $\pm$ 3 \\

% \end{tabular}
% % \vspace{-0.4cm}
% % \end{adjustbox}
% \end{table}

\begin{table*}[t]
% \normalsize
% \footnotesize
\caption{Perplexity Showing Sample Efficiency Across All Methods for Known Users. Lower is Better.}
\label{tab:known_results}

    \centering
    \resizebox{\textwidth}{!}{
    \begin{tabular}{lc|ccccc}
    
& \# Samples & FedPC         & FedPC (Frozen)            & FedPC (Generated)            & Split-Learning & Meta-Learning             \\
    
  \hline
 \parbox[t]{0mm}{\multirow{4}{*}{\rotatebox[origin=c]{90}{Reddit}}} &&&&&& \\

&1            &  219.5 $\pm$ 35.7                & 146.2 $\pm$ 2.3       &  \textbf{120.2 $\pm$ 1.4}      & 1297.5 $\pm$ 21.9  & 226.2 $\pm$ 3.7 \\
&5             & 131.6 $\pm$ 10.1                 & 136.9 $\pm$ 3.4            & \textbf{123.3 $\pm$ 2.8}    &  994.3 $\pm$ 27.8   & 234.7 $\pm$ 5.1 \\
&15             & \textbf{111.4 $\pm$ 3.5}       & 132.6 $\pm$ 4.7            & 120.0 $\pm$ 3.3    & 691.3 $\pm$ 34.3   & 227.1 $\pm$ 8.4 \\
&All             & 189.5 $\pm$ 6.7                & 167.9 $\pm$ 2.0         & \textbf{124.9 $\pm$ 1.3}    & 930.4 $\pm$ 30.9   & 241.4 $\pm$ 2.1 \\
\hline \hline
 \parbox[t]{0mm}{\multirow{5}{*}{\rotatebox[origin=c]{90}{TV Shows}}} &&&&&&\\
&1            &  57.2 $\pm$ 3.6                & \textbf{50.3 $\pm$ 1.6}        &  51.6 $\pm$ 1.3     & 359.4 $\pm$ 28.2  & 111.7 $\pm$ 4.6 \\
&5             & 51.5$\pm$ 1.5                & \textbf{50.7 $\pm$ 2.1}            & 51.7 $\pm$ 2.0   & 244.5 $\pm$ 15.1   & 110.0 $\pm$ 6.5 \\
&15             & \textbf{48.8 $\pm$ 1.7}                & 51.0 $\pm$ 2.1            & 51.7 $\pm$ 2.0    & 167.7 $\pm$ 8.6   & 111.9 $\pm$ 6.1 \\
&All             & \textbf{46.7 $\pm$ 1.7}                & 51.2 $\pm$ 2.0            & 52.1 $\pm$ 2.6    & 82.1 $\pm$ 3.3   & 113.0 $\pm$ 4.7 \\
    \end{tabular}}
\end{table*}

\subsection{Datasets}
We conduct our experiments using two datasets, a smaller dataset of TV Show scripts (``Friends'' \citep{chen2016character} and ``Game of Thrones'' \citep{got_data}) and a larger dataset of Reddit posts \citep{convokit}. Each dataset has a diverse set of individuals as well as clearly defined contexts/styles (i.e., TV shows or subreddits). These properties enable us to not only compare our approach to baseline approaches for personalized predictions, but they also enable us to move users between contexts or styles (e.g., producing text for a ``Friends'' character under a ``Game of Thrones'' context). By generating sequences for different users under new styles, we demonstrate the power of FedPC for personal, stylized prediction. Additional information about the datasets used in this work is given in the appendix.

%Statistics for each dataset are given in Table \ref{table:datasets}. 
% The TV Show dataset is constructed by merging scripts from two shows, ``Friends'' and ``Game of Thrones.'' \change{We use ConvoKit \citep{convokit} to gather the ``Friends'' Corpus \citep{chen2016character}, and retain the six main characters. We use a set of ``Game of Thrones'' scripts \citep{got_data} to query for the thirteen characters with the highest utterance-count.
% Our merged dataset has 19 characters, 60650 utterances, and two contexts.}
% % The result is a dataset with two highly distinct styles and topics of conversation, with a set of personalities that are very closely tied to their respective contexts (``Friends'' or ``Game of Thrones''). 
% The average utterance count for each character is 3370, with ``Friends'' characters having more utterances than ``Game of Thrones'' characters.
% %This dynamic creates an imbalance in the dataset, where the majority of utterances are drawn from the ``Friends'' context, but the majority of characters are drawn from the ``Game of Thrones'' dataset.

% Our Reddit experiments use the ``reddit-corpus-small`` dataset from ConvoKit \citep{convokit}, which includes posts from the top-100 subreddits over a set period of time. We filter the dataset to only include users with at least 50 utterances and contexts (subreddits) with at least 150 utterances. 
% The resulting dataset has 326 characters, 30260 utterances, and 57 contexts.

% This filtering removed users and contexts without enough data for a proper train/test split. 
For both datasets, we treat each sentence from a speaker (i.e., TV Show character or Reddit user) as an independent utterance and we only consider utterances with at least three tokens. For experiments on known users, we perform a 60/20/20 Train/Validation/Test data split. For experiments on novel, unseen users, we perform a 70/15/15 split of Reddit users, and we manually select the ``Friends'' and ``Game of Thones'' users to include in each data fold. For both sets of experiments, all contexts are seen during training.

\begin{table*}[t]
% \normalsize
% \footnotesize
\caption{Perplexity Showing Sample Efficiency Across All Methods for Withheld Users. Lower is Better}
\vspace{2mm}
\label{tab:withheld_results}

    \centering
    \resizebox{\textwidth}{!}{
    \begin{tabular}{lc|ccccc}
    
& \# Samples & FedPC         & FedPC (Frozen)            & FedPC (Generated)            & Split-Learning & Meta-Learning             \\
    
  \hline
 \parbox[t]{0mm}{\multirow{4}{*}{\rotatebox[origin=c]{90}{Reddit}}} &&&&&& \\

& 1            &  594.3 $\pm$ 973.8                & 202.0 $\pm$ 5.9        &  \textbf{117.3 $\pm$ 1.8}     & 922.9 $\pm$ 27.8  & 213.9 $\pm$ 6.0 \\
& 5             & 139.4 $\pm$ 4.4                & 202.9 $\pm$ 10.9            & \textbf{117.5 $\pm$ 2.7}    & 655.9 $\pm$ 18.8   & 212.2 $\pm$ 5.4 \\
& 15             & 117.4 $\pm$ 1.9                & 203.6 $\pm$ 11.2            & \textbf{116.6 $\pm$ 2.6}    & 449.2 $\pm$ 11.4   & 211.7 $\pm$ 3.7 \\
& All             & \textbf{101.1$\pm$ 2.2}                & 202.2 $\pm$ 7.6            & 117.9 $\pm$ 2.8    & 309.3 $\pm$ 8.3   & 212.8 $\pm$ 5.2 \\
\hline \hline
 \parbox[t]{0mm}{\multirow{5}{*}{\rotatebox[origin=c]{90}{TV Shows}}} &&&&&&\\
& 1            &  205.1 $\pm$ 292.2                & 96.4 $\pm$ 10.4       &  \textbf{68.7 $\pm$ 5.9}     & 283.6 $\pm$ 30.9  & 113.5 $\pm$ 13.1 \\
& 5             & 68.6$\pm$ 5.6               & 90.1 $\pm$ 4.9            & \textbf{66.7 $\pm$ 6.3}    & 220.7 $\pm$ 29.2   & 111.4 $\pm$ 13.3 \\
& 15             & \textbf{62.1 $\pm$ 5.0}                & 97.6 $\pm$ 6.8            & 66.1 $\pm$ 5.5    & 158.1 $\pm$ 20.0   & 117.3 $\pm$ 10.5 \\
& All             & \textbf{52.3 $\pm$ 3.3}                & 98.2 $\pm$ 9.5            & 68.6 $\pm$ 5.1    & 96.7 $\pm$ 14.5   & 114.2 $\pm$ 17.0\\
    
    \end{tabular}}
\end{table*}
\subsection{Results and Discussion}
\label{subsec:results}
All experiments are repeated fifteen times, and means and standard deviations for performance and runtime results are presented in Tables \ref{tab:known_results}, \ref{tab:withheld_results}, and \ref{table:runtime_results}. 
Tables \ref{tab:known_results} and \ref{tab:withheld_results} show that our approach is able to generate sensible language for both held-out user instances and known users. 
Both embedding-based approaches presented in this paper (i.e., FedPC with generated or learned embeddings) show drastic improvements over baselines in terms of both sample- and runtime-efficiency, and are more suitable for real-world on-device language models.

\paragraph{Summary} With known users, FedPC achieves perplexity as low as 46.7 and 100.3, on the TV Show and Reddit datasets, respectively, compared to the best baseline perplexities of 82.1 and 233.2 (a 45-50\% improvement). For unknown users, FedPC achieves perplexities of 52.3 and 97.6, respectively, compared to baselines at 96.7 and 212.7 (a 45-55\% improvement). FedPC training times are between 25-400\% faster than baseline training times. Finally, FedPC uses 0.001\% of the memory that baseline methods use for stylized personalization.

\paragraph{Memory Costs} FedPC incurs a significantly lower memory cost than prior Split-Learning based approaches \citep{li2019fedmd,collins2021fedrep,dinh2020pfedme,tang2021pfedkm,rudovic2021personalized,gupta2018split}.
The Split-Learning baselines require maintaining a model-head for each user and context present in the dataset, and the size of these model heads is proportional to the size of the vocabulary.
On each client-device, a user's personal model head and all context heads need to be stored in memory and used in forward passes.
In our work, every GPT model head is approximately 154 MB (being $768 \times 50257$ parameters).
To update the model on-device, one would need to store a model head corresponding to every possible context. Our Reddit dataset involves 57 contexts, totalling an additional $\sim$ 8GB of data in memory. This memory requirement for personalized heads could become infeasible for real-world tasks, particularly for on-device inference or backpropagation on mobile devices. 
Using FedPC, which only requires the addition of a drastically smaller preference embedding, the total amount of memory required on device to store the embeddings is only $\sim$171 KB (0.001\% of the memory required by separate model heads).

\begin{table*}
\caption{Training and Testing run-time for FedPC and our baselines, in milliseconds. Lower is better.}
\vspace{2mm}
% The global network is frozen for all baselines except for ``None'', where the entire model is updated on-device.}
\label{table:runtime_results}
% \normalsize
\centering
\begin{tabular}{l|ccccc}
Method & FedPC         & FedPC (Frozen)            & FedPC (Generated)            & Split-Learning & Meta-Learning             \\
%\hline
%Perplexity    & 72.54\% $\pm$ 1.31\% & 93.90\% $\pm$ .20\% & 93.52\% $\pm$ .45\% & 74.52\% $\pm$ 2.68\% \\
\hline \hline
Train Pass Time            &  88.18 $\pm$ 24.104                & \textbf{43.57 $\pm$ 11.99}        &  55.96 $\pm$ 12.41     & 222.08 $\pm$ 37.55  & 111.81 $\pm$ 22.33 \\
Test Pass Time             & 40.37 $\pm$ 11.76                & 40.25 $\pm$ 12.10           & 47.02 $\pm$ 12.63    & 65.42 $\pm$ 16.49   & \textbf{36.77 $\pm$ 8.95}
\end{tabular}
\end{table*}

\paragraph{Sample Efficiency}
FedPC is able to outperform Split-Learning and Meta-Learning models with significantly fewer samples across both experiments and both datasets.
This trend is reflected regardless of whether embeddings are generated or learned through backpropagation. 
When embeddings are learned, FedPC improves with online data to more effectively model the given user's style as more data is made available to the model. 
Conversely, while the generated embeddings exhibit greater sample performance with a single sample, they are unable to improve with more data. 
For both known and with-held users, FedPC with generated embeddings is unable to effectively update the preference embedding to improve generation performance.
Finally, we see an increase in perplexity for Reddit users with all available data when using FedPC. This result suggests that it is possible to \textit{overfit} preference embeddings, as we see an increase in perplexity from 15 to ``All'' samples (Table \ref{tab:known_results}).

We observe no improvement for the Meta-Learning baseline, regardless of how much data is available for each user. This lack of improvement suggests that the model is not capable of rapidly personalizing to a single user or context with only a handful of available samples. Only updating the model head may be insufficient when the base, shared model head must generalize across all possible contexts and characters.

The Split-Learning baseline, on the other hand, does show significant improvement with increasing amounts of data for withheld and known users. In our known user experiments, all personal model heads 
% (i.e., associated and used for only a single character or Reddit user) 
should have already been well-tuned to personal preferences. Our result therefore suggests that context-specific model heads are over-generalized to their respective contexts, and must be refined to better-align with individual users.

\begin{table*}[t]
    \caption{Generated Examples using Arya, from ``Game of Thrones'' (GoT) and Chandler, from ``Friends''.}
    \vspace{2mm}
    \label{tab:qualitative}
    \centering
    \resizebox{\textwidth}{!}{
    \begin{tabular}{c|c|l|l}
        Character & Show & ``We Must'' & ``I think'' \\
        \hline \hline
        \multirow{3}{*}{Chandler} & \multirow{3}{*}{Friends} & be careful! I'm not going to get a divorce. & I'll be able to do this.\\
         & & be a little bit more relaxed than we're here. & I'm a good man \\
          & & be the one who's the one who's the one... & I'm a big fan of you\\
        \hline
        \multirow{3}{*}{Chandler} & \multirow{3}{*}{GoT} & be honest with you. & I'm going to be a little more serious\\
         & & be very nervous about the possibility of a bomb attack. & I'm going to be a little bit of a jerk \\
         & & be a little nervous about the situation & I'm going to have a big secret.\\
        \hline
        \multirow{3}{*}{Arya} & \multirow{3}{*}{GoT} & be a little more careful. & you can't help me \\
         & & be careful about the dangers of the sea. & of the people\\
          & & be wary of the possibility of a coup. & you're not going to be a hero?\\
        \hline
        \multirow{3}{*}{Arya} & \multirow{3}{*}{Friends} & be a thief & I'm not a bad person \\
         & & be a hero. & I can do it\\
         & & be a little girl. & I should have a chance to do something\\
    \end{tabular}}
\end{table*}

\paragraph{Runtime Efficiency}
FedPC incurs significantly lower training costs than both Split-Learning and Meta-Learning approaches to personalization. 
% As FedPC only requires updating or generating a personal or context embedding.
While Meta-Learning baseline does not have the memory-constraints of the split-learning model in terms of storing \textit{additional} model heads, training the Meta-Learning baseline still involves computing gradients over all $768 \times 50257$ parameters in the shared output layer. As we show in Table~\ref{table:runtime_results}, this leads to a significantly more costly training time for each user. Similarly, the Split-Learning baseline must update \textit{at least} two model heads for each backward pass, requiring gradient computation for $2 \times 768 \times 50257$ parameters. If a user is active in multiple contexts, then additional context model-heads must be used, further exacerbating the training cost of the Split-Learning approach. The Split-Learning approach must also leverage these additional context model-heads at test-time, resulting in the slowest forward-passes of any baseline.

In contrast to prior approaches, training for FedPC only requires updating $2 \times 768$ parameters.
% (one embedding for personal preferences, and one for context preferences)
This reduced computation results in significantly lower training times. When we train an embedding-generator, there is an increase in training times reflecting the added cost of computing gradients for the embedding generator. Additionally, there is a test-time penalty incurred by the added forward-pass parameters. When running inference with any version of FedPC, preference embeddings are combined and then prepended to the input utterance. This process results in marginally slower test times with FedPC relative to the Meta-Learning baseline.

\paragraph{Qualitative Results}
Our qualitative results in Table \ref{tab:qualitative} demonstrate the power of FedPC. Not only is our model able to complete sequences for a character in their ``home'' context (i.e., the context from which all of their data is drawn), but we are also able to stylize generation for characters, bringing them into \textit{new} contexts. We present generated samples from a ``Game of Thrones'' (GoT) character (Arya) with a ``Friends'' context embedding and a GoT context embedding. We see that Arya's generated sequences are distinct under the two different contexts. Under the GoT context, Arya's utterances match the theme of the show, suggesting danger and revolution. Under the ``Friends'' context, Arya's utterances change to instead reflect more mundane, modern language while still preserving personal attributes of the character.

Across all of our experiments, particularly the novel experimental evaluation on held-out user-instances, our results provide evidence that embedding-conditioned personalization within federated learning can be effectively applied to real-world use-cases. 
FedPC offers a promising avenue of future work towards on-device language models, capable of efficient language generation with respect to compute-power and data availability.% \todo{@Andrew, we could potentially remove this entire paragraph if we need to. Or just add one of these experiments to the end of the first paragraph.}

\section{Limitations}
Firstly, although our embedding-generator offers a promising avenue of personalizing without any on-device gradient computation, our generator is currently unable to improve on its generated embeddings given more examples for a given user. 
%Our results (Sec. \ref{subsec:results}) show that generated preference embeddings perform well, but we do not see improvement with more data.
As shown in our results from Sec~\ref{subsec:results}, while the model can generate an effective preference embedding for a user with a single sample, it is unable to improve with more data. 
In future work, we hope to explore approaches to facilitate a generator which can effectively modify embeddings given additional data. 

Secondly, our approach caters to confidentiality by ensuring that user-data and embeddings remain on-device, however we have not incorporated differential privacy in our experiments \citep{li2020secure}. 
%Gaussian privacy is achieved when a user's gradients with respect to the global model are retained on device, ensuring that each individual user's model parameters cannot be recreated by utilizing their gradients.
% Currently our model sends gradients, based on on-device user data, back to the server in order to perform an average update of the global model. 
Future work may apply differential privacy to guarantee user privacy while personalizing and contributing feature encoder information to a central server. %enable our approach to function with the addition of gaussian noise to the gradients such that user-privacy is preserved. 
Finally, it is important to note that FedPC does not solve all problems within the scope of language generation models. As FedPC offers a path forward to facilitate privacy protection and efficient on-device learning for large language models, future work may extend FedPC to additional problems (e.g., language summarization or turn-based dialogue generation).

%\todo{@Matthew - You mentioned something about addressing the future of dialog systems here. We weren't sure what you meant. We were never strictly looking to solve problems related to improving dialog systems, as we haven't compared to any dialog baselines or included that as a part of our contributions. We're worried that speaking about ``dialog''-related limitations opens us up for criticism, especially given the venue is EMNLP. Do you just want us to specify that our system doesn't solve all existing problems in the language generation space?}

\section{Conclusion}
We present FedPC, a new approach to personalized federated learning, enabling efficient and high-performance personalization to client devices by leveraging preference embeddings. 
Combining context with individual personal preferences, FedPC outperforms baselines even when allotted a lower computational budget.
We also provide a method of generating preference embeddings through inference alone, providing personalization with no on-device gradient computation, and we show comparable performance to FedPC with learned embeddings.
We presented experiments on two datasets, TV Show scripts and Reddit user data, presenting empirical evidence of the utility of FedPC towards personalizing to unseen users in a federated learning setting, i.e. a 50\% improvement in terms of runtime to fine-tune on new users as well as perplexity.
We also demonstrated qualitative results, showing the power of separate personal and context embeddings and enabling stylization of users in new contexts.
%Our experiments show improvement in terms of both data and performance compared to prior state-of-the-art federated learning baselines like personalized-heads or Fed-Avg.
Our results show that FedPC offers a promising path forward for personalization within federated learning, achieving superior quantitative results, and requiring significantly less training time relative to baseline approaches.

\section{Acknowledgements}
This work was supported by the Office of Naval Research (ONR) under award N00014-19-1-2076.
Andrew Silva was supported by the Apple Scholars in AI/ML PhD fellowship. 
\bibliographystyle{unsrtnat}
\bibliography{references}  %%% Uncomment this line and comment out the ``thebibliography'' section below to use the external .bib file (using bibtex) .

\begin{thebibliography}{44}
\providecommand{\natexlab}[1]{#1}
\providecommand{\url}[1]{\texttt{#1}}
\expandafter\ifx\csname urlstyle\endcsname\relax
  \providecommand{\doi}[1]{doi: #1}\else
  \providecommand{\doi}{doi: \begingroup \urlstyle{rm}\Url}\fi

\bibitem[Dudy et~al.(2021)Dudy, Bedrick, and Webber]{dudy2021refocusing}
Shiran Dudy, Steven Bedrick, and Bonnie Webber.
\newblock Refocusing on relevance: Personalization in nlg.
\newblock \emph{arXiv preprint arXiv:2109.05140}, 2021.

\bibitem[Li et~al.(2020{\natexlab{a}})Li, Chang, and Chi]{li2020secure}
Yiwei Li, Tsung-Hui Chang, and Chong-Yung Chi.
\newblock Secure federated averaging algorithm with differential privacy.
\newblock In \emph{2020 IEEE 30th International Workshop on Machine Learning
  for Signal Processing (MLSP)}, pages 1--6. IEEE, 2020{\natexlab{a}}.

\bibitem[McMahan et~al.(2017)McMahan, Moore, Ramage, Hampson, and
  y~Arcas]{mcmahan2017communication}
Brendan McMahan, Eider Moore, Daniel Ramage, Seth Hampson, and Blaise~Aguera
  y~Arcas.
\newblock Communication-efficient learning of deep networks from decentralized
  data.
\newblock In \emph{Artificial intelligence and statistics}, pages 1273--1282.
  PMLR, 2017.

\bibitem[Jones(1999)]{jones1999automatic}
Karen~Sparck Jones.
\newblock Automatic summarizing: factors immarizing: factors and directions.
\newblock \emph{Advances in automatic text summarization}, page~1, 1999.

\bibitem[Wu et~al.(2021)Wu, Ma, and Yang]{wu2021personalized}
Yuwei Wu, Xuezhe Ma, and Diyi Yang.
\newblock Personalized response generation via generative split memory network.
\newblock In \emph{Proceedings of the 2021 Conference of the North American
  Chapter of the Association for Computational Linguistics: Human Language
  Technologies}, pages 1956--1970, 2021.

\bibitem[Zhang et~al.(2018)Zhang, Dinan, Urbanek, Szlam, Kiela, and
  Weston]{zhang2018personalizing}
Saizheng Zhang, Emily Dinan, Jack Urbanek, Arthur Szlam, Douwe Kiela, and Jason
  Weston.
\newblock Personalizing dialogue agents: I have a dog, do you have pets too?
\newblock \emph{arXiv preprint arXiv:1801.07243}, 2018.

\bibitem[Cheng et~al.(2019)Cheng, Fang, and Ostendorf]{cheng2019dynamic}
Hao Cheng, Hao Fang, and Mari Ostendorf.
\newblock A dynamic speaker model for conversational interactions.
\newblock In \emph{Proceedings of the 2019 Conference of the North American
  Chapter of the Association for Computational Linguistics: Human Language
  Technologies, Volume 1 (Long and Short Papers)}, pages 2772--2785, 2019.

\bibitem[Lin et~al.(2019{\natexlab{a}})Lin, Madotto, Shin, Xu, and
  Fung]{lin2019moel}
Zhaojiang Lin, Andrea Madotto, Jamin Shin, Peng Xu, and Pascale Fung.
\newblock Moel: Mixture of empathetic listeners.
\newblock \emph{arXiv preprint arXiv:1908.07687}, 2019{\natexlab{a}}.

\bibitem[Gupta and Raskar(2018)]{gupta2018split}
Otkrist Gupta and Ramesh Raskar.
\newblock Distributed learning of deep neural network over multiple agents.
\newblock \emph{Journal of Network and Computer Applications}, 116:\penalty0
  1--8, 2018.

\bibitem[Finn et~al.(2017)Finn, Abbeel, and Levine]{finn2017maml}
Chelsea Finn, Pieter Abbeel, and Sergey Levine.
\newblock Model-agnostic meta-learning for fast adaptation of deep networks.
\newblock In Doina Precup and Yee~Whye Teh, editors, \emph{Proceedings of the
  34th International Conference on Machine Learning}, volume~70 of
  \emph{Proceedings of Machine Learning Research}, pages 1126--1135, 06--11 Aug
  2017.

\bibitem[Nichol et~al.(2018)Nichol, Achiam, and Schulman]{nichol2018reptile}
Alex Nichol, Joshua Achiam, and John Schulman.
\newblock On first-order meta-learning algorithms.
\newblock \emph{arXiv preprint arXiv:1803.02999}, 2018.

\bibitem[Collins et~al.(2021)Collins, Hassani, Mokhtari, and
  Shakkottai]{collins2021fedrep}
Liam Collins, Hamed Hassani, Aryan Mokhtari, and Sanjay Shakkottai.
\newblock Exploiting shared representations for personalized federated
  learning.
\newblock In Marina Meila and Tong Zhang, editors, \emph{Proceedings of the
  38th International Conference on Machine Learning}, volume 139 of
  \emph{Proceedings of Machine Learning Research}, pages 2089--2099, 18--24 Jul
  2021.

\bibitem[Dinh et~al.(2020)Dinh, Tran, and Nguyen]{dinh2020pfedme}
Canh~T Dinh, Nguyen~H Tran, and Tuan~Dung Nguyen.
\newblock Personalized federated learning with moreau envelopes.
\newblock \emph{arXiv preprint arXiv:2006.08848}, 2020.

\bibitem[Li et~al.(2020{\natexlab{b}})Li, Sahu, Zaheer, Sanjabi, Talwalkar, and
  Smith]{li2020federated}
Tian Li, Anit~Kumar Sahu, Manzil Zaheer, Maziar Sanjabi, Ameet Talwalkar, and
  Virginia Smith.
\newblock Federated optimization in heterogeneous networks.
\newblock \emph{Proceedings of Machine Learning and Systems}, 2:\penalty0
  429--450, 2020{\natexlab{b}}.

\bibitem[Li and Wang(2019)]{li2019fedmd}
Daliang Li and Junpu Wang.
\newblock Fedmd: Heterogenous federated learning via model distillation.
\newblock \emph{arXiv preprint arXiv:1910.03581}, 2019.

\bibitem[Arivazhagan et~al.(2019)Arivazhagan, Aggarwal, Singh, and
  Choudhary]{arivazhagan2019federated}
Manoj~Ghuhan Arivazhagan, Vinay Aggarwal, Aaditya~Kumar Singh, and Sunav
  Choudhary.
\newblock Federated learning with personalization layers.
\newblock \emph{arXiv preprint arXiv:1912.00818}, 2019.

\bibitem[Kim et~al.(2021)Kim, Park, Jung, and Yoo]{kim2021spatio}
Joongheon Kim, Seunghoon Park, Soyi Jung, and Seehwan Yoo.
\newblock Spatio-temporal split learning.
\newblock In \emph{2021 51st Annual IEEE/IFIP International Conference on
  Dependable Systems and Networks-Supplemental Volume (DSN-S)}, pages 11--12.
  IEEE, 2021.

\bibitem[Rudovic et~al.(2021)Rudovic, Tobis, Kaltwang, Schuller, Rueckert,
  Cohn, and Picard]{rudovic2021personalized}
Ognjen Rudovic, Nicolas Tobis, Sebastian Kaltwang, Bj{\"o}rn Schuller, Daniel
  Rueckert, Jeffrey~F Cohn, and Rosalind~W Picard.
\newblock Personalized federated deep learning for pain estimation from face
  images.
\newblock \emph{arXiv preprint arXiv:2101.04800}, 2021.

\bibitem[Paulik et~al.(2021)Paulik, Seigel, Mason, Telaar, Kluivers, van Dalen,
  Lau, Carlson, Granqvist, Vandevelde, et~al.]{paulik2021federated}
Matthias Paulik, Matt Seigel, Henry Mason, Dominic Telaar, Joris Kluivers,
  Rogier van Dalen, Chi~Wai Lau, Luke Carlson, Filip Granqvist, Chris
  Vandevelde, et~al.
\newblock Federated evaluation and tuning for on-device personalization: System
  design \& applications.
\newblock \emph{arXiv preprint arXiv:2102.08503}, 2021.

\bibitem[Tang et~al.(2021)Tang, Guo, and Guo]{tang2021pfedkm}
Xueyang Tang, Song Guo, and Jingcai Guo.
\newblock Personalized federated learning with clustered generalization.
\newblock \emph{arXiv preprint arXiv:2106.13044}, 2021.

\bibitem[Silva et~al.(2022)Silva, Metcalf, Apostoloff, and
  Theobald]{silva2022fedembed}
Andrew Silva, Katherine Metcalf, Nicholas Apostoloff, and Barry-John Theobald.
\newblock Fedembed: Personalized private federated learning.
\newblock \emph{arXiv preprint arXiv:2202.09472}, 2022.

\bibitem[Jiang et~al.(2019)Jiang, Kone{\v{c}}n{\`y}, Rush, and
  Kannan]{jiang2019improving}
Yihan Jiang, Jakub Kone{\v{c}}n{\`y}, Keith Rush, and Sreeram Kannan.
\newblock Improving federated learning personalization via model agnostic meta
  learning.
\newblock \emph{arXiv preprint arXiv:1909.12488}, 2019.

\bibitem[Fallah et~al.(2020)Fallah, Mokhtari, and
  Ozdaglar]{fallah2020personalized}
Alireza Fallah, Aryan Mokhtari, and Asuman Ozdaglar.
\newblock Personalized federated learning: A meta-learning approach.
\newblock \emph{arXiv preprint arXiv:2002.07948}, 2020.

\bibitem[Deng et~al.(2020)Deng, Kamani, and Mahdavi]{deng2020apfl}
Yuyang Deng, Mohammad~Mahdi Kamani, and Mehrdad Mahdavi.
\newblock Adaptive personalized federated learning.
\newblock \emph{arXiv preprint arXiv:2003.13461}, 2020.

\bibitem[Hanzely and Richt{\'a}rik(2020)]{hanzely2020federated}
Filip Hanzely and Peter Richt{\'a}rik.
\newblock Federated learning of a mixture of global and local models.
\newblock \emph{arXiv preprint arXiv:2002.05516}, 2020.

\bibitem[Hanzely et~al.(2020)Hanzely, Hanzely, Horv{\'a}th, and
  Richt{\'a}rik]{hanzely2020lower}
Filip Hanzely, Slavom{\'\i}r Hanzely, Samuel Horv{\'a}th, and Peter
  Richt{\'a}rik.
\newblock Lower bounds and optimal algorithms for personalized federated
  learning.
\newblock \emph{arXiv preprint arXiv:2010.02372}, 2020.

\bibitem[Lin et~al.(2019{\natexlab{b}})Lin, Madotto, Wu, and
  Fung]{lin2019personalizing}
Zhaojiang Lin, Andrea Madotto, Chien-Sheng Wu, and Pascale Fung.
\newblock Personalizing dialogue agents via meta-learning.
\newblock \emph{arXiv preprint arXiv:1905.10033}, 2019{\natexlab{b}}.

\bibitem[Tamar et~al.(2018)Tamar, Rohanimanesh, Chow, Vigorito, Goodrich,
  Kahane, and Pridmore]{tamar2018imitation}
Aviv Tamar, Khashayar Rohanimanesh, Yinlam Chow, Chris Vigorito, Ben Goodrich,
  Michael Kahane, and Derik Pridmore.
\newblock Imitation learning from visual data with multiple intentions.
\newblock In \emph{International Conference on Learning Representations}, 2018.

\bibitem[Hsiao et~al.(2019)Hsiao, Kuo, and Sun]{hsiao2019learning}
Fang-I Hsiao, Jui-Hsuan Kuo, and Min Sun.
\newblock Learning a multi-modal policy via imitating demonstrations with mixed
  behaviors.
\newblock \emph{arXiv preprint arXiv:1903.10304}, 2019.

\bibitem[Paleja et~al.(2020)Paleja, Silva, Chen, and
  Gombolay]{paleja2020interpretable}
Rohan Paleja, Andrew Silva, Letian Chen, and Matthew Gombolay.
\newblock Interpretable and personalized apprenticeship scheduling: {L}earning
  interpretable scheduling policies from heterogeneous user demonstrations.
\newblock In H.~Larochelle, M.~Ranzato, R.~Hadsell, M.~F. Balcan, and H.~Lin,
  editors, \emph{Advances in Neural Information Processing Systems}, volume~33,
  pages 6417--6428. Curran Associates, Inc., 2020.

\bibitem[Schrum et~al.(2022)Schrum, Hedlund-Botti, Moorman, and
  Gombolay]{schrum2022mind}
Mariah~L Schrum, Erin Hedlund-Botti, Nina Moorman, and Matthew~C Gombolay.
\newblock Mind meld: Personalized meta-learning for robot-centric imitation
  learning.
\newblock In \emph{Proceedings of the 2022 ACM/IEEE International Conference on
  Human-Robot Interaction}, pages 157--165, 2022.

\bibitem[Lu et~al.(2021)Lu, Huang, Zhan, and Zhuang]{lu2021federated}
Yujie Lu, Chao Huang, Huanli Zhan, and Yong Zhuang.
\newblock Federated natural language generation for personalized dialogue
  system.
\newblock \emph{arXiv preprint arXiv:2110.06419}, 2021.

\bibitem[Yang and Flek(2021)]{yang2021towards}
Diyi Yang and Lucie Flek.
\newblock Towards user-centric text-to-text generation: A survey.
\newblock In \emph{International Conference on Text, Speech, and Dialogue},
  pages 3--22. Springer, 2021.

\bibitem[Majumder et~al.(2020)Majumder, Jhamtani, Berg-Kirkpatrick, and
  McAuley]{majumder2020like}
Bodhisattwa~Prasad Majumder, Harsh Jhamtani, Taylor Berg-Kirkpatrick, and
  Julian McAuley.
\newblock Like hiking? you probably enjoy nature: Persona-grounded dialog with
  commonsense expansions.
\newblock \emph{arXiv preprint arXiv:2010.03205}, 2020.

\bibitem[Kim et~al.(2020)Kim, Kim, and Kim]{kim2020will}
Hyunwoo Kim, Byeongchang Kim, and Gunhee Kim.
\newblock Will i sound like me? improving persona consistency in dialogues
  through pragmatic self-consciousness.
\newblock \emph{arXiv preprint arXiv:2004.05816}, 2020.

\bibitem[Li and Liang(2021)]{li-liang-2021-prefix}
Xiang~Lisa Li and Percy Liang.
\newblock Prefix-tuning: Optimizing continuous prompts for generation.
\newblock In \emph{Proceedings of the 59th Annual Meeting of the Association
  for Computational Linguistics and the 11th International Joint Conference on
  Natural Language Processing (Volume 1: Long Papers)}, pages 4582--4597,
  Online, August 2021. Association for Computational Linguistics.
\newblock \doi{10.18653/v1/2021.acl-long.353}.

\bibitem[Ha et~al.(2016)Ha, Dai, and Le]{ha2016hypernetworks}
David Ha, Andrew Dai, and Quoc~V Le.
\newblock Hypernetworks.
\newblock \emph{arXiv preprint arXiv:1609.09106}, 2016.

\bibitem[Shamsian et~al.(2021)Shamsian, Navon, Fetaya, and
  Chechik]{shamsian2021personalized}
Aviv Shamsian, Aviv Navon, Ethan Fetaya, and Gal Chechik.
\newblock Personalized federated learning using hypernetworks.
\newblock In \emph{International Conference on Machine Learning}, pages
  9489--9502. PMLR, 2021.

\bibitem[Vaswani et~al.(2017)Vaswani, Shazeer, Parmar, Uszkoreit, Jones, Gomez,
  Kaiser, and Polosukhin]{vaswani2017attention}
Ashish Vaswani, Noam Shazeer, Niki Parmar, Jakob Uszkoreit, Llion Jones,
  Aidan~N Gomez, {\L}ukasz Kaiser, and Illia Polosukhin.
\newblock Attention is all you need.
\newblock \emph{Advances in neural information processing systems}, 30, 2017.

\bibitem[Wolf et~al.(2019)Wolf, Debut, Sanh, Chaumond, Delangue, Moi, Cistac,
  Rault, Louf, Funtowicz, et~al.]{wolf2019huggingface}
Thomas Wolf, Lysandre Debut, Victor Sanh, Julien Chaumond, Clement Delangue,
  Anthony Moi, Pierric Cistac, Tim Rault, R{\'e}mi Louf, Morgan Funtowicz,
  et~al.
\newblock Huggingface's transformers: State-of-the-art natural language
  processing.
\newblock \emph{arXiv preprint arXiv:1910.03771}, 2019.

\bibitem[Chen and Choi(2016)]{chen2016character}
Yu-Hsin Chen and Jinho~D Choi.
\newblock Character identification on multiparty conversation: Identifying
  mentions of characters in tv shows.
\newblock In \emph{Proceedings of the 17th Annual Meeting of the Special
  Interest Group on Discourse and Dialogue}, pages 90--100, 2016.

\bibitem[Koirala(2019)]{got_data}
Shekhar Koirala.
\newblock Game\_of\_thrones.
\newblock \url{github.com/shekharkoirala/Game\_of\_Thrones}, 2019.

\bibitem[Chang et~al.(2020)Chang, Chiam, Fu, Wang, Zhang, and
  Danescu{-}Niculescu{-}Mizil]{convokit}
Jonathan~P. Chang, Caleb Chiam, Liye Fu, Andrew~Z. Wang, Justine Zhang, and
  Cristian Danescu{-}Niculescu{-}Mizil.
\newblock Convokit: {A} toolkit for the analysis of conversations.
\newblock In Olivier Pietquin, Smaranda Muresan, Vivian Chen, Casey Kennington,
  David Vandyke, Nina Dethlefs, Koji Inoue, Erik Ekstedt, and Stefan Ultes,
  editors, \emph{Proceedings of the 21th Annual Meeting of the Special Interest
  Group on Discourse and Dialogue, SIGdial 2020, 1st virtual meeting, July 1-3,
  2020}, pages 57--60. Association for Computational Linguistics, 2020.
\newblock URL \url{https://aclanthology.org/2020.sigdial-1.8/}.

\bibitem[Kingma and Ba(2014)]{kingma2014adam}
Diederik~P Kingma and Jimmy Ba.
\newblock Adam: A method for stochastic optimization.
\newblock \emph{arXiv preprint arXiv:1412.6980}, 2014.

\end{thebibliography}

\appendix
\section{Generation Algorithm}
\label{sec:appendix}

At each time-step during inference, the embeddings are updated by the following equations. 
\begin{gather*} 
e_t = Multi\_Head\_Attn(W_{<t}, LN(e_{t-1}), W_{<t}) \\
    e_t = LN(LN(e_t) + LN(e_{t-1})) \\
    \label{eq:gen_eq}
    e_t = LN(FFN(e_t) + LN(e_{t-1}) 
\end{gather*}

For the first timestep, $e_{t-1}$ is initialized as $\phi$ or $\psi$ for personal and context embeddings respectively, and $LN$ represents a layer normalization function. We apply future-masking to prevent any future-information in the sequence from leaking forward into the rest of the model. After processing the entire utterance, the generated embedding is updated to the final value of $e_t$, which can then be stored on-device for future processing. 
An updated algorithm which applies the generator to predict preference embeddings can be found in Alg~\ref{alg:appendix_training_loop}.
% While this approach is specific to the language-modeling domain, analogous methods may be devised for other domains.

\begin{algorithm*}
\caption{Personalized Federated Learning Loop with Generated Embeddings}
\label{alg:appendix_training_loop}
\begin{algorithmic}[1]
\STATE {\bfseries Given:} Training objective, $\mathcal{L}$,  Client devices $D$
\STATE {\bfseries Given:} Number of client steps, $K$
\STATE {\bfseries Given:} Number of global steps, $N$
\STATE {\bfseries Initialize:} Global model, $\theta$, Context embeddings $\phi$, Context Generator $\Gamma$, Client Generator $\nu$
\STATE{\bfseries Initialize:} Personal embeddings on-device $\psi$
\FOR{$n \in N$}
    \FOR{$d \in D$}
        \STATE $\theta_d = \theta, \phi_d = \phi, \Gamma_d = \Gamma, \nu_d = \nu$
        \FOR{$k \in K$}
            \STATE Sample $B_d$ from client's on-device data
            \STATE $\theta_d \leftarrow \theta_d + \nabla_\theta\mathcal{L}(\theta_d, \phi_{d,c}, \psi_d, B_d)$ // Fine-tune global model with local data
            \STATE $\phi_d \leftarrow \nu_d(\theta_d, \phi_{d,c}, B_d)$ // Generate context embedding from local data
            \STATE $\psi_d \leftarrow \Gamma_{d}(\theta_d, \psi_d, B_d)$ // Generate personal embedding from local data
            \STATE $\nu_d \leftarrow \nu_d + \nabla_\nu \mathcal{L}(\theta_d, \phi_{d,c}, \psi_d, B_d)$ // Update client Generator
            \STATE $\Gamma_d \leftarrow \Gamma_d + \nabla_\Gamma \mathcal{L}(\theta_d, \phi_{d,c}, \psi_d, B_d)$ // Update context Generator
        \ENDFOR
        \STATE $\nabla_{\theta_d} \leftarrow \theta-\theta_d$ // compute final client $\theta$ gradients
        \STATE $\nabla_{\Gamma_d} \leftarrow \Gamma - \Gamma_d$ // compute final client $\Gamma$ gradients
        \STATE $\nabla_{\nu_d} \leftarrow \nu - \nu_d$ // compute final client $\nu$ gradients
        \STATE Return $\nabla_{\theta_d}$, $\nabla_{\nu_d}$ and $\nabla_{\Gamma_d}$ to the server
    \ENDFOR
    \STATE $\nabla_\theta \leftarrow \frac{1}{D}  \sum_d^D \nabla_{\theta_d} $  //  calculate average $\theta$ gradients
    \STATE $\nabla_\Gamma \leftarrow \frac{1}{D}  \sum_d^D \nabla_{\Gamma_d} $ //  calculate average $\Gamma$ gradients
    \STATE $\nabla_\nu \leftarrow \frac{1}{D}  \sum_d^D \nabla_{\nu_d} $ //  calculate average $\nu$ gradients
    \STATE $\theta \leftarrow \theta + \nabla_{\theta}$
    \STATE $\phi \leftarrow \phi + \nabla_{\phi}$
\ENDFOR
\end{algorithmic}
\end{algorithm*}
\section{Training Details}
All models are initialized with the DistilGPT2 pre-trained model from Huggingface \citep{wolf2019huggingface}. All layers of the model are frozen, and FedPC only backpropagates error to personal and context preference embeddings. For our Meta-Learning baseline, the last layer is unfrozen and all users jointly update this final output layer (note: there is no dedicated context head in this approach). Our Split-Learning baseline assigns a unique model head to each user and to each context, and each user only updates their own model head and the contexts that they use. 

All models are trained for 55 epochs over their training datasets using the Adam optimizer \citep{kingma2014adam} for global updates (learning rate = 1) and local updates (learning rate = 0.001). Each client (character or Reddit user) makes 10 local updates before passing their pooled gradient information back to the server. During training, each client samples 15 data points per training pass. For local fine-tuning updates at test-time, each user makes 15 updates using a small portion of the test data (the data used for fine-tuning is not used for testing).

All models use a frozen DistilGPT2 model from HuggingFace as their initialization. After empirical experimentation, we opted to freeze the majority of the DistilGPT2 parameters by default. This freezing helped to save on computational and memory costs as well as improving generalization performance across diverse users. As a result of this freezing, shared learning and personalization updates will only affect model heads, shared embeddings, and/or personal embeddings. 

FedPC leverages a standard federated averaging training procedure (FedAvg) \citep{mcmahan2017communication} with the addition of a FedProx penalty term \citep{li2020federated} to regularize on-device client updates back to the globally-averaged model. Empirically, FedProx improved performance for all methods. We fix the FedProx $\mu$ parameter to 1.

Training was carried out on an NVIDIA A40 GPU with 48GB of memory. Due to limitations of the GPU, not all context-heads could be stored in memory at once for our Split Learning baseline when working with the Reddit dataset. The GPU could only accommodate 14 model heads in addition to the DistilGPT2 model, but the dataset featured 57 unique subreddits. To work around this limitation, 13 context heads were active at all times, and the parameters of those heads were saved and overwritten as necessary to ensure that each user had access to their required context heads.

\section{Dataset Information}
The TV Show dataset is constructed by merging scripts from two shows, ``Friends'' and ``Game of Thrones.'' We use ConvoKit \citep{convokit} to gather the ``Friends'' Corpus \citep{chen2016character}, and retain the six main characters. We use a set of ``Game of Thrones'' scripts \citep{got_data} to query for the thirteen characters with the highest utterance-count.
Our merged dataset has 19 characters, 60650 utterances, and two contexts.
% The result is a dataset with two highly distinct styles and topics of conversation, with a set of personalities that are very closely tied to their respective contexts (``Friends'' or ``Game of Thrones''). 
The average utterance count for each character is 3370, with ``Friends'' characters having more utterances than ``Game of Thrones'' characters.
%This dynamic creates an imbalance in the dataset, where the majority of utterances are drawn from the ``Friends'' context, but the majority of characters are drawn from the ``Game of Thrones'' dataset.

Our Reddit experiments use the ``reddit-corpus-small`` dataset from ConvoKit \citep{convokit}, which includes posts from the top-100 subreddits over a set period of time. We filter the dataset to only include users with at least 50 utterances and contexts (subreddits) with at least 150 utterances. 
The resulting dataset has 326 characters, 30260 utterances, and 57 contexts.

\end{document}